\documentclass[
twocolumn,
]{ceurart}

\usepackage{listings}
\usepackage{epsfig}
\usepackage{amsmath}
\usepackage{amssymb}
\usepackage{booktabs}
\usepackage{multirow,bm}
\usepackage{adjustbox}
\usepackage{makecell}
\usepackage{rotating}
\usepackage{tablefootnote}
\usepackage[para,online,flushleft]{threeparttable}
\usepackage{array}
\usepackage{subfigure}
\usepackage{xcolor}
\usepackage{pythonhighlight}
\usepackage{hyperref}
\usepackage{colortbl}

\usepackage{algorithm}
\usepackage{algpseudocode}
\usepackage{mathtools}

\usepackage{lipsum}
\usepackage{color}
\usepackage{multirow}

\definecolor{Yize_color}{rgb}{0.9, 0.1, 0.8}

\definecolor{mygray}{gray}{.8}
\definecolor{mygray2}{gray}{.65}

\begin{document}

\copyrightclause{Copyright for this paper by its authors. Use permitted under Creative Commons License Attribution 4.0 International (CC BY 4.0).}
\copyrightyear{2023}

\conference{The AAAI-23 Workshop on Artificial Intelligence Safety (SafeAI 2023), Feb 13-14, 2023, Washington, D.C., US}

\title{Less is More: Data Pruning for Faster Adversarial Training}



\author[1]{Yize Li}[%
email=li.yize@northeastern.edu,
]
\fnmark[1]
\author[1]{Pu Zhao}[%
email=p.zhao@northeastern.edu,
]
\author[1]{Xue Lin}[%
email=xue.lin@northeastern.edu,
]
\address[1]{Northeastern University, 360 Huntington Ave, Boston, MA 02115}

\author[2]{Bhavya Kailkhura}[%
email=kailkhura1@llnl.gov,
]
\author[2]{Ryan Goldhahn}[%
email=goldhahn1@llnl.gov,
]
\address[2]{Lawrence Livermore National Laboratory, 7000 East Ave, Livermore, CA 94550}

\fntext[1]{Corresponding author.}

\begin{abstract}
Deep neural networks (DNNs) are sensitive to adversarial examples, resulting in fragile and unreliable performance in the real world. Although adversarial training (AT) is currently one of the most effective methodologies to robustify DNNs, it is computationally very expensive ({\it e.g.}, $5\sim10\times$ costlier than standard training). To address this challenge, existing approaches focus on single-step AT, referred to as Fast AT, reducing the overhead of adversarial example generation. Unfortunately, these approaches are known to fail against stronger adversaries. 
To make AT computationally efficient without compromising robustness, this paper takes a different view of the efficient AT problem. Specifically, we propose to minimize redundancies at the data level by leveraging \emph{{data pruning}}. Extensive experiments demonstrate that the data pruning based AT can achieve similar or superior robust (and clean) accuracy as its unpruned counterparts while being significantly faster. For instance, proposed strategies accelerate CIFAR-10 training up to $3.44\times$ and CIFAR-100 training to $2.02\times$. Additionally, the data pruning methods can readily be reconciled with existing adversarial acceleration tricks to obtain the striking speed-ups of $5.66\times$ and $5.12\times$ on CIFAR-10, $3.67\times$ and $3.07\times$ on CIFAR-100 with TRADES and MART, respectively. 
\end{abstract}


\maketitle

\section{Introduction}
Deep neural networks (DNNs) achieve great success in various machine learning tasks, such as image classification~\cite{9156610, 50870}, object detection \cite{8627998, ZAIDI2022103514}, language modeling \cite{NIPS2017_3f5ee243, pmlr-v162-borgeaud22a} and so on.  
However, the reliability and security concerns of DNNs limit their wide deployment in real-world applications. 
For example, imperceptible perturbations added to inputs by adversaries (known as adversarial examples) \cite{chen2017zoo, ijcai2018p543, NEURIPS2020_11f38f8e} can cause incorrect predictions during inference. Therefore, many research efforts are devoted to designing robust DNNs against adversarial examples \cite{pmlr-v80-athalye18a,pmlr-v80-wong18a, NEURIPS2019}.

Adversarial Training (AT)~\cite{madry2017towards} is one of the most effective defense approaches to improving adversarial robustness. AT is formulated as a min-max problem, with the inner maximization aiming to generate adversarial examples, and the outer minimization aiming to train a model based on them. 
However, to achieve better defense with higher robustness, the iterative AT is required to generate stronger adversarial examples with more steps in the inner problem, leading to expensive computation costs. 
In response to this difficulty, a number of approaches investigate efficient AT, such as Fast AT \cite{wong2020fast} and their variants \cite{9157154, kim2021understanding} via single-step adversarial attacks. Unfortunately, these cheaper training approaches are known to attain poor performance on stronger adversaries and suffer from `catastrophic overfitting' \cite{wong2020fast, andriushchenko2020understanding}, where Projected Gradient Descent (PGD) robustness is gained at the beginning, but later the robust accuracy  decreases to 0 suddenly. In this regard, there does not seem to exist a satisfactory solution to achieve optimal robustness with moderate computation cost.

In this paper, we propose to overcome the above limitation by exploring a new perspective---leveraging \emph{\textbf{data pruning}} during AT. 
Differing from the prior Fast AT-based solutions that focus on 
the AT algorithm, 
we attain efficiency by selecting the representative subset of training samples and performing AT on this smaller dataset. 


Although several recent works explore data pruning for efficient standard training (see \cite{bartoldson2022compute} for a survey), data pruning for efficient AT is not well investigated. 
To the best of our knowledge, the most relevant one is \cite{https://doi.org/10.48550/arxiv.2207.00694}, which speeds up AT by the loss-based data pruning. However, the random sub-sampling  outperforms their data pruning scheme
in terms of clean accuracy, robustness, and training efficiency, raising doubts about the feasibility of the proposed approach. 
In contrast, we propose to perform data pruning in two ways: 1) by maximizing the log-likelihood of the subset on the validation dataset, and 2) by minimizing the gradient disparity between the subset and the full dataset. We implement these approaches with two AT objectives: TRADES \cite{zhang2019theoretically} and MART \cite{Wang2020Improving}. Experimental results show that we can achieve training acceleration up to $3.44\times$ on CIFAR-10 and $2.02\times$ on CIFAR-100. In addition, incorporating our proposed data pruning with Bullet-Train \cite{hua2021bullettrain}, which allocates dynamic computing cost to categorized training data, further improves the speed-ups by $5.66\times$ and $3.67\times$ on CIFAR-10 and CIFAR-100, respectively. 
Our main contributions are summarized below. 
\begin{itemize}
    \item We explore efficient AT from the lens of data pruning, where the acceleration is achieved by only focusing on the representative subset of the data. 
    \item We propose two data pruning algorithms, Adv-GRAD-MATCH and Adv-GLISTER, and perform a comprehensive experimental study. We demonstrate that our data pruning methods yield consistent effectiveness across diverse robustness evaluations, {\it e.g.}, PGD \cite{madry2017towards} and AutoAttack \cite{croce2020reliable}.
    \item Furthermore, combining our efficient AT framework with the existing Bullet-Train approach \cite{hua2021bullettrain} achieves state-of-the-art performance in training cost.
\end{itemize}

\section{Related Work}
\paragraph{Adversarial attacks and defenses.}
Adversarial attacks \cite{madry2017towards, Goodfellow2015explaining,carlini2017towards, Croce_2019_ICCV, Zhang2022BIA} refer to detrimental techniques that inject imperceptible perturbations into the inputs and mislead decision making process of networks. 
In this paper, we mainly investigate $\ell_p$ attacks, where $p \in\{ 0, 1,2,\infty\}$. 
Fast Gradient Sign Method (FGSM) \cite{Goodfellow2015explaining} is the cheapest one-shot adversarial attack. Basic Iterative Method (BIM) \cite{https://doi.org/10.48550/arxiv.1607.02533}, 
Projected Gradient
Descent (PGD) \cite{madry2017towards} and CW \cite{carlini2017towards} are stronger attacks that are iterative in nature.
Adversarial examples are used for the assessment of model robustness.  AutoAttack \cite{croce2020reliable} ensembles multiple attack strategies to perform a fair and reliable evaluation of adversarial robustness.

Various defense methods \cite{10.1145/3133956.3134057, Liao2018CVPR, Mustafa_2019_ICCV, gong2022reverse} have been proposed to tackle the vulnerability of DNNs against adversarial examples, while most of the approaches are built over AT, where perturbed inputs are fed to DNNs to learn from adversarial examples. Projected Gradient Descent (PGD) based AT is one of the most popular defense strategies \cite{madry2017towards}, which uses a multi-step adversary. Training only with adversarial samples can lead to a drop in clean accuracy \cite{Su_2018_ECCV}. To improve the trade-off between accuracy and robustness, TRADES \cite{zhang2019theoretically} and MART \cite{Wang2020Improving} compose the training loss with both the natural error term and the robustness regularization term. 
Curriculum Adversarial Training (CAT) \cite{ijcai2018p520} robustifies DNNs by adjusting PGD steps arranging from weak attack strength to strong attack strength, while Friendly Adversarial Training (FAT) \cite{pmlr-v119-zhang20z} performs early-stopped PGD for 
adversarial examples. 

\paragraph{Efficient adversarial training.}
Despite PGD-based training showing empirical robustness against adversarial examples, the learning overhead is usually dramatically larger than the standard training, {\it e.g.}, $5\sim10\times$ computation consumption depending on the number of steps used in generating adversarial examples. The major work to achieve training efficiency focuses on how to reduce the number of attack steps and maintain the stability of one-step FGSM-based AT. Free AT \cite{shafahi2019adversarial} performs FGSM perturbations and updates model weights on the simultaneous mini-batch. FAST AT \cite{wong2020fast} generates FGSM attacks with random initialization but still suffers from `catastrophic overfitting'. Therefore, Gradient alignment regularization \cite{andriushchenko2020understanding}, suitable inner interval (step size) for the adversarial direction \cite{kim2021understanding}, and Fast Bi-level AT (FAST-BAT) \cite{zhangrevisiting} are proposed to prevent such failure.

\paragraph{Data pruning.}
Efficient learning through data subset selection economizes on training resources. Proxy functions \cite{coleman2020selection, Kaushal_2019_WACV} take advantage of the feature representation from the tiny proxy model to select the most informative subset for training the larger one. Coreset-based algorithms \cite{Feldman2020} mine for a small representative subset that approximates the entire dataset following established criteria. CRAIG \cite{pmlr-v119-mirzasoleiman20a} selects the training data subset which approximates the full gradient and GRAD-MATCH \cite{pmlr-v139-killamsetty21a} minimizes the gradient matching error. GLISTER \cite{Killamsetty_Sivasubramanian_Ramakrishnan_Iyer_2021} prunes the training data by maximizing log-likelihood for the validation set.

\section{Data Pruning Based Adversarial Training}
\label{sec:formulation}

\subsection{Preliminaries}
AT \cite{madry2017towards} aims to solve the min-max optimization problem  as follows:
\begin{align}
\min _{\theta} \frac{1}{|D|}\sum_{(x, y) \in \mathcal{D}}\left[\max _{\delta \in \triangle} \mathcal{L}(\theta; x+\delta, y)\right],
\label{eq: adv}
\end{align}
where $\theta$ is the model parameter, $x$ and $y$ denote the data sample and label from the training dataset $\mathcal{D}$, $\delta$ denotes imperceptible adversarial perturbations injected into $x$ under the norm constraint by the constant strength $\epsilon$, {\it i.e.}, ${\triangle} :=\{ \|\delta\|_{\infty} \leq \epsilon \}$, and $\mathcal{L}$ is the training loss. During the adversarial procedure, the optimization first maximizes the inner approximation for adversarial attacks and then minimizes the outer training error over the model parameter $\theta$. 
A typical adversarial example generation procedure
involves multiple steps for the stronger adversary, {\it e.g.},
\begin{align}
x^{t+1} & = \operatorname{Proj}_{\triangle}\left(x^{t}+\alpha \operatorname{sign}\left(\nabla_{x^{t}} \mathcal{L}\left(\theta; x^{t}, y\right)\right)\right),
\label{eq: adv_gen}
\end{align}
where the projection follows $\epsilon$-ball at the  step $t$ with step size $\alpha$, using the sign of gradients.

\subsection{General Formulation for Adversarial Data Pruning}
Our adversarial data pruning consists of two steps: adversarial subset selection and AT with the subset of data. In the specified epoch, adversarial subset selection first finds a representative subset of data from the entire training dataset. Next, AT is performed with the selected subset. Though the size of the subset keeps the same in different iterations, the data in the subset is updated in each iteration based on the different status of the model weights. 
We formulate the AT with the data subset in Eq.~\eqref{eq: subadv} and adversarial subset selection in Eq.~\eqref{eq: subadv2}.
\begin{align}
\min _{\theta} \frac{1}{k}\sum_{(x,y)\in \mathcal{S}}\left[\max _{\delta \in \triangle} \mathcal{L}(\theta; x+\delta, y)\right],
\label{eq: subadv}
\end{align}
\begin{align}
\underset{\mathcal{S} \subseteq \mathcal{D},|\mathcal{S}|=k}{\operatorname{min}}  {{G}(\mathcal{S})}
\label{eq: subadv2}
\end{align}
where $\mathcal{D}$ represents the complete training set and $\delta$ represents the perturbation under ${ l_{\infty} }$ norm constraint ${\triangle}$. The selected subset $\mathcal{S}$ with the size $k$ is obtained by optimizing the function $G$, which aims to narrow the difference between $\mathcal{D}$ and $\mathcal{S}$ under specific criteria with model parameters $\theta$.  
Note that the data selection step is performed periodically to achieve computational savings.
\begin{figure*}[h]
  \centering
  \subfigure[TRADES.]{\includegraphics[width=0.35\linewidth]{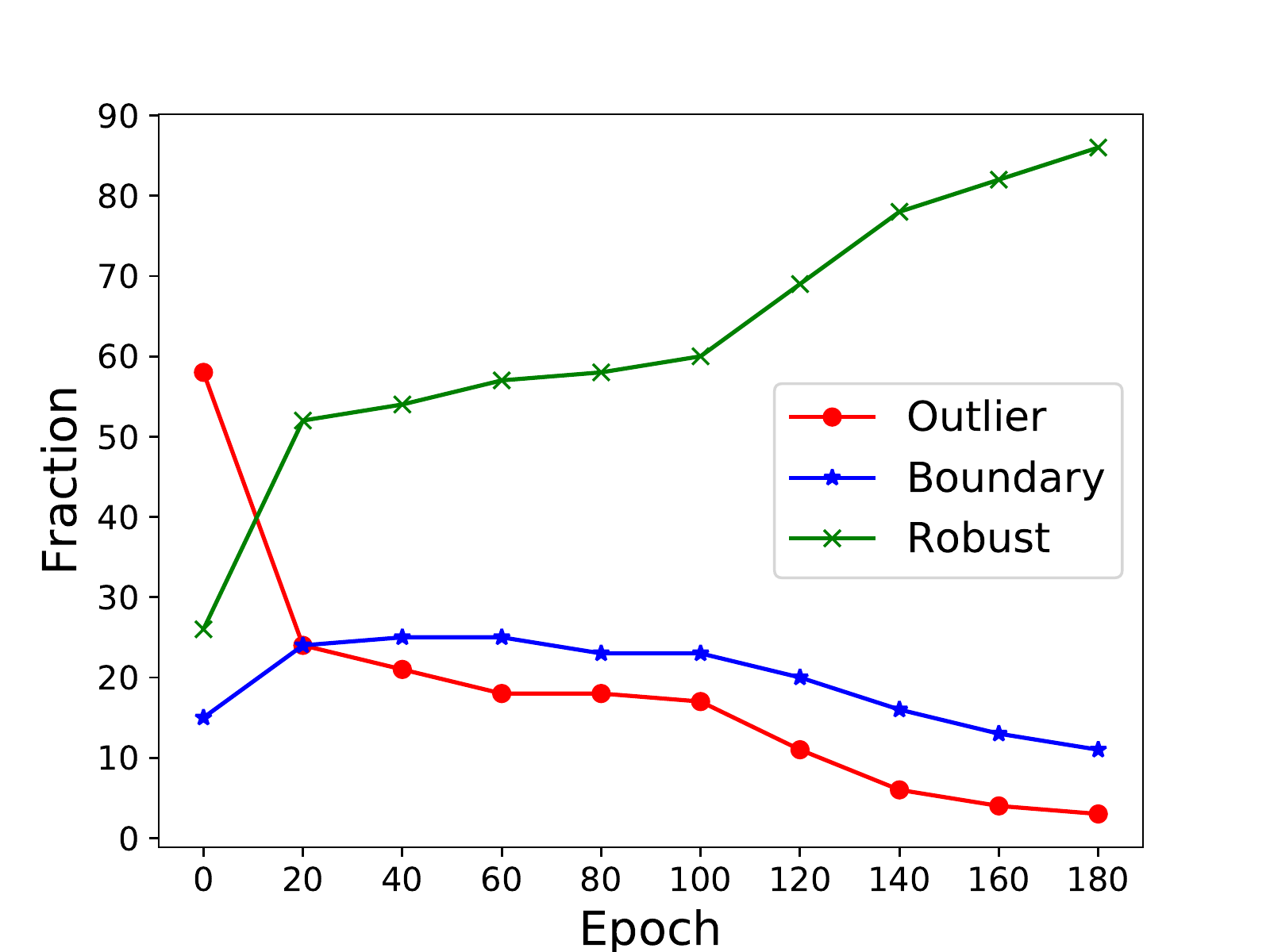}}\hspace{-5mm}\vspace{-1mm}
  \subfigure[Bullet.]{\includegraphics[width=0.35\linewidth]{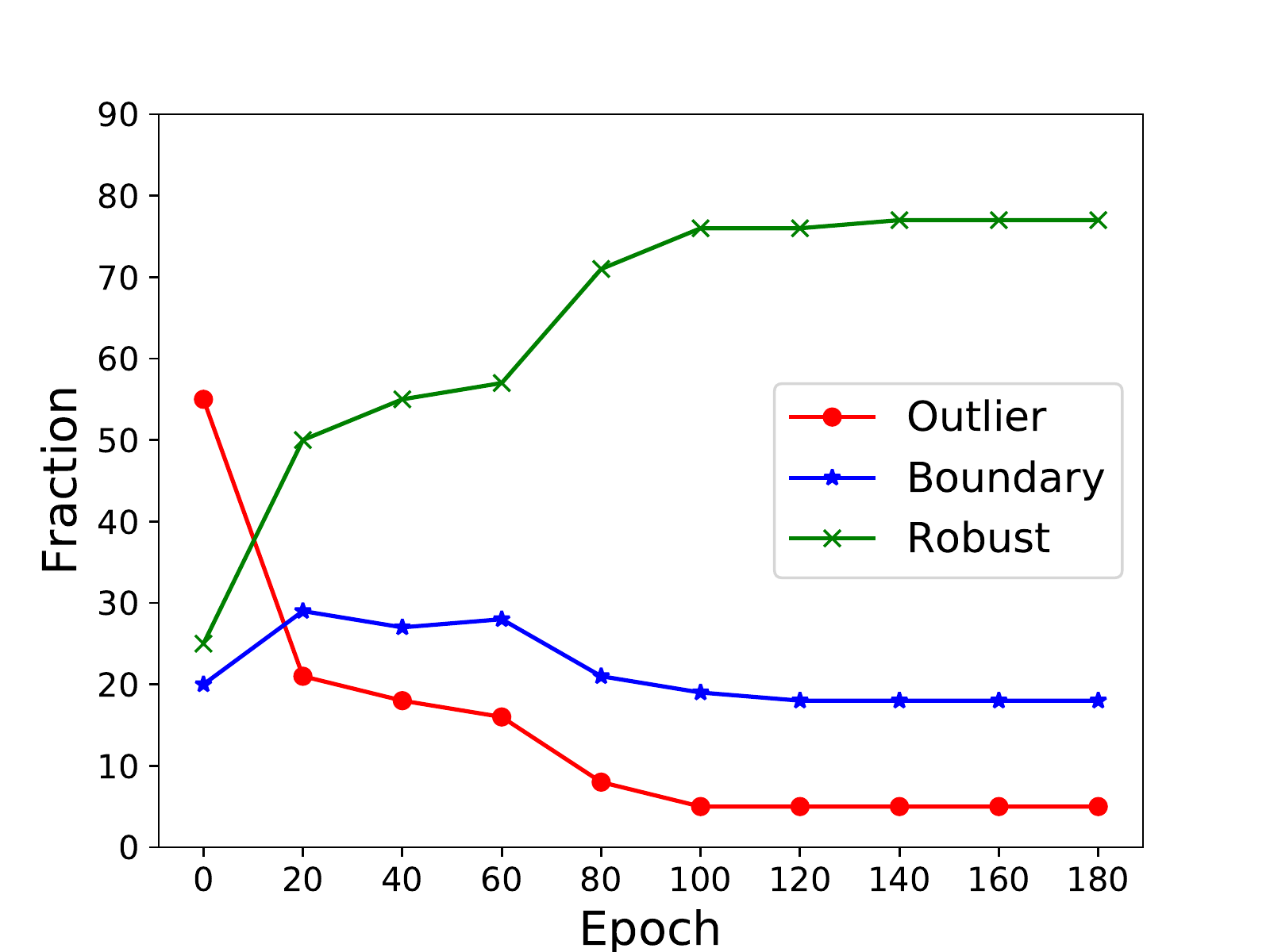}}\hspace{-5mm}\vspace{-1mm}
  \subfigure[Adv-GRAD-MATCH.]{\includegraphics[width=0.35\linewidth]{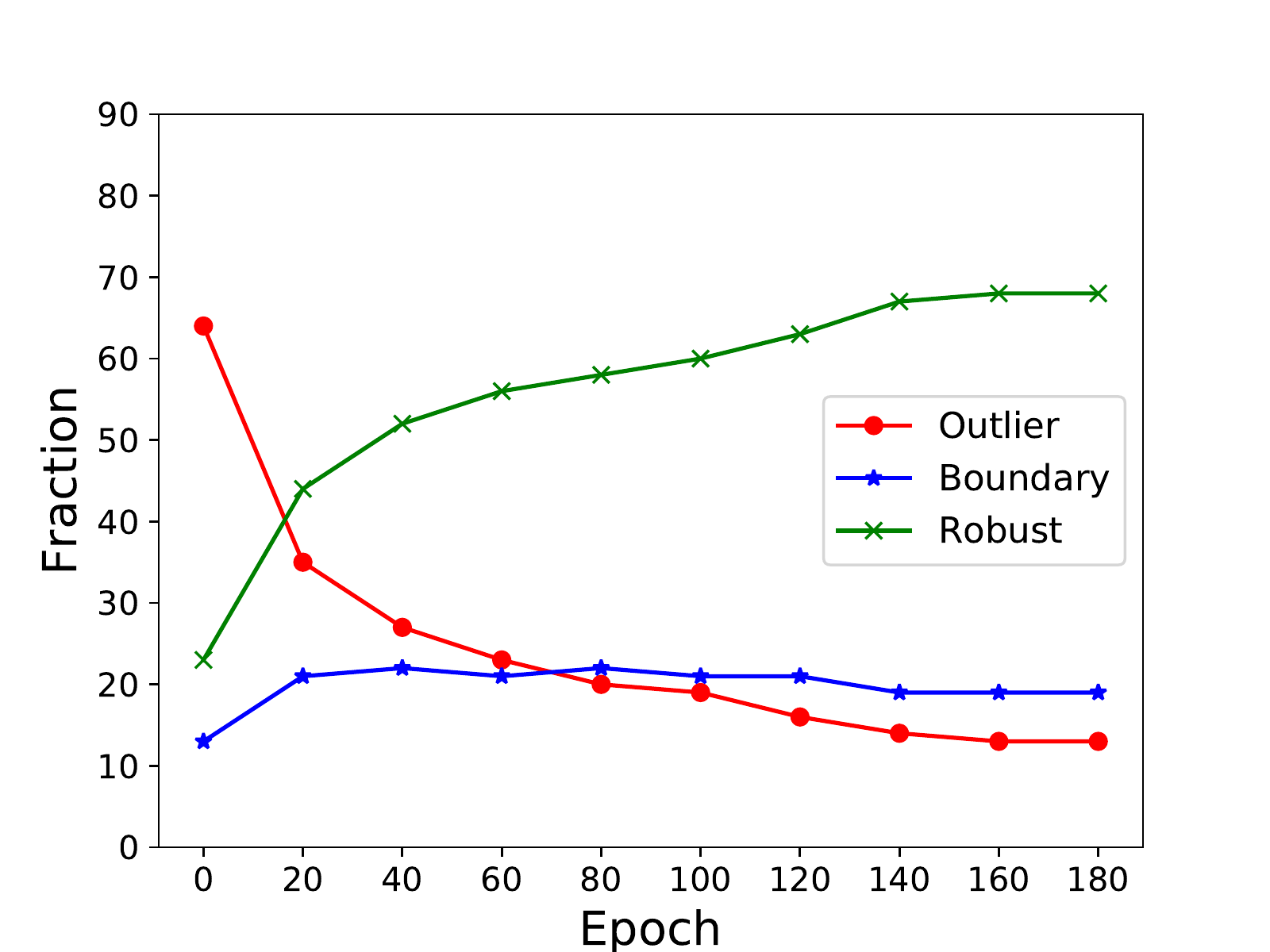}}\hspace{-5mm}\vspace{-1mm}
  \subfigure[Adv-GLISTER.]{\includegraphics[width=0.35\linewidth]{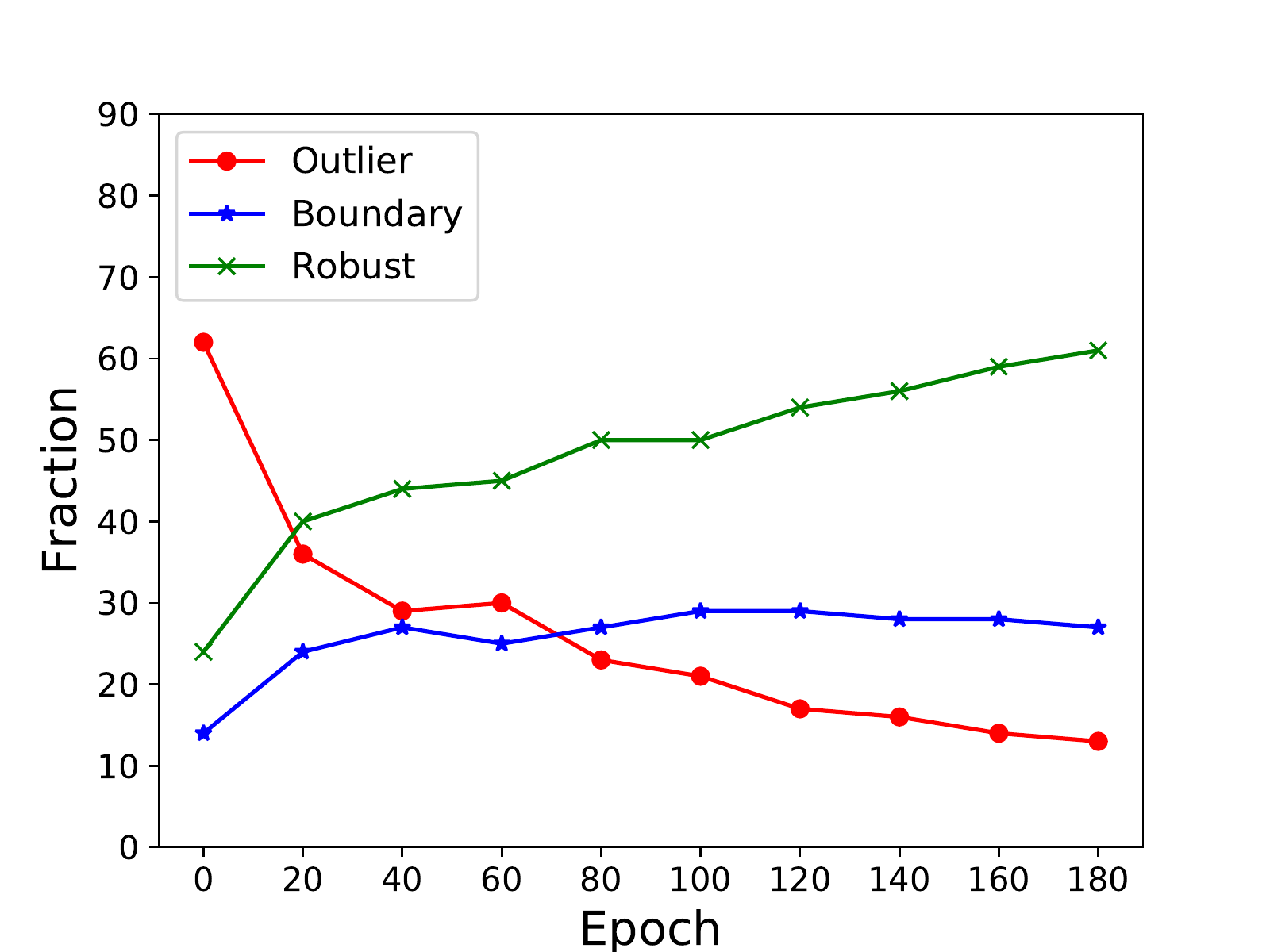}}\hspace{-5mm}\vspace{-1mm}
  \subfigure[Adv-GRAD-MATCH\&Bullet.]{\includegraphics[width=0.35\linewidth]{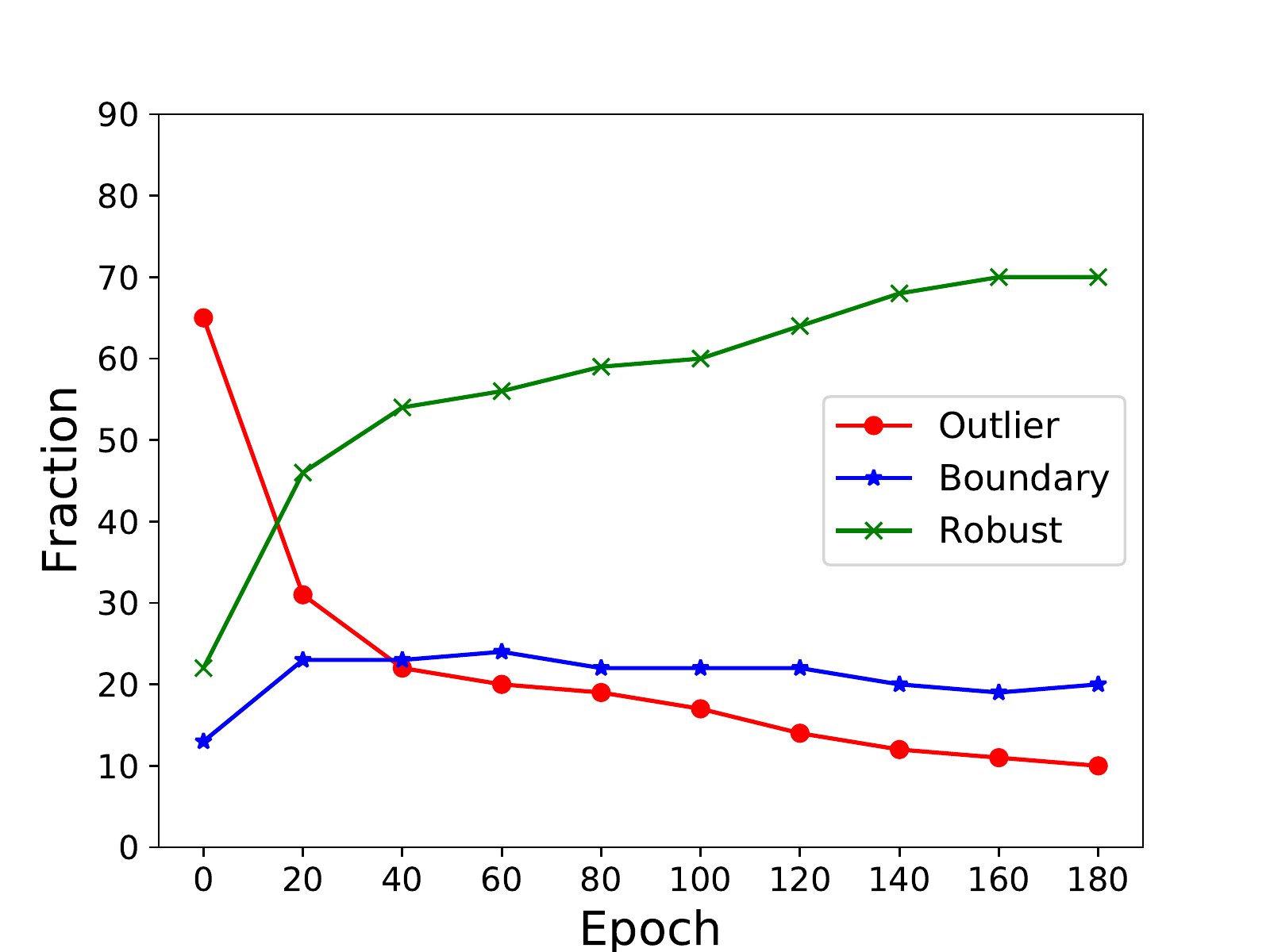}}\hspace{-5mm}\vspace{-1mm}
  \subfigure[Adv-GLISTER\&Bullet.]{\includegraphics[width=0.35\linewidth]{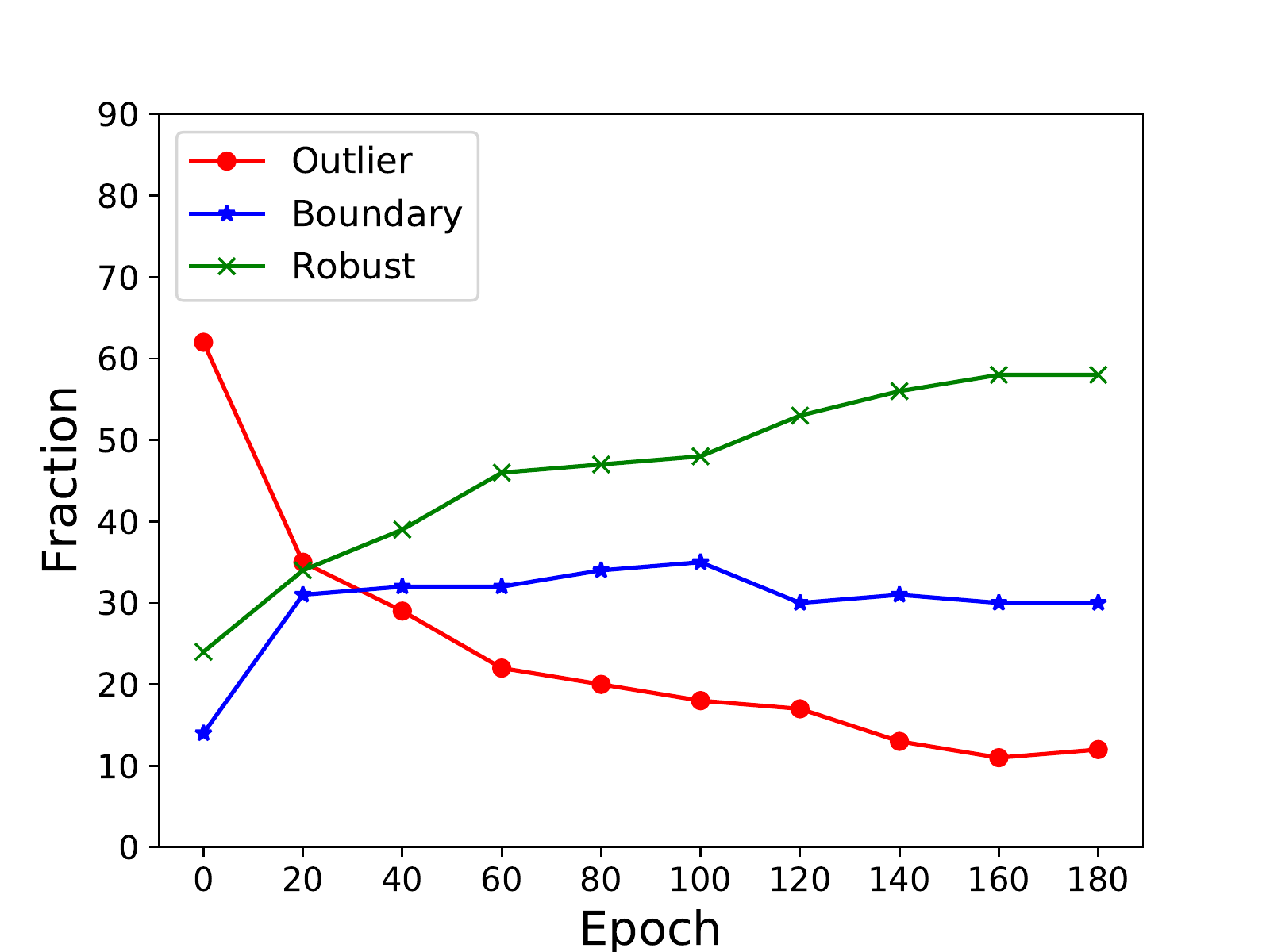}}\hspace{-5mm}\vspace{-1mm}
  \caption{\footnotesize{Tracking of adversarial robustness during 200 epochs of training. \textcolor{red}{Red}, \textcolor{green}{Green} and \textcolor{blue}{Blue} denote outlier, robust and boundary examples, respectively.}}
    \label{fig:tracking}  
  \end{figure*}

  \begin{table*}[htbp]
  \caption{TRADES results where data pruning methods use only 30\% data points on CIFAR-10 and 50\% data points on CIFAR-100 for 100 epochs of training.}
  \label{tab:TRADES}
  \centering
  \scalebox{0.9}{
  \begin{tabular}{c|ccccccc}
  \toprule
  \hline
 \multirow{2}{*}{Dataset} & \multirow{2}{*}{Method}    & \multirow{2}{*}{Clean} & \multicolumn{3}{c}{PGD}                                      &  \multirow{2}{*}{AutoAttack} &  \multirow{2}{*}{Time/epoch (Speed-up)} \\ 
  &                         &                    & $4/255$     & $8/255$     & $16/255$    &                            &                            \\ \hline
  \multirow{6}{*}{CIFAR-10} 
  & TRADES  \cite{zhang2019theoretically}           &82.73 & 69.17 & 51.83 & 19.43 & 49.06 & 416.20 (-)    \\
  & Bullet \cite{hua2021bullettrain}    & 84.60	& {70.24}	& 50.82	& 16.05	& 47.93	& 193.06 (2.16$\times$)
                 \\ 
  & Adv-GLISTER (Ours)   & 77.62	& 63.06	& 46.06	& 16.52	& 41.61	& 120.70 (3.45$\times$)
                \\ 
  & Adv-GRAD-MATCH (Ours) & 75.67	& 61.85	& 45.96	& 17.49	& 42.19	& 138.19 (3.01$\times$)
                \\ 
  & Adv-GLISTER\&Bullet (Ours)   & 79.21	& 63.02	& 44.52	& 13.33	& 40.77	& \textbf{72.91 (5.66$\times$)}
       \\ 
  &Adv-GRAD-MATCH\&Bullet (Ours) & 77.57	& 62.00	& 45.13	& 14.65	& 41.94	& 87.38 (4.76$\times$)
      \\ \hline
  \multirow{6}{*}{CIFAR-100} 
  & TRADES \cite{zhang2019theoretically}  &55.85	& 40.31	& 27.35	& {10.71}	& 23.39	& 387.72 (-)
     \\
  & Bullet \cite{hua2021bullettrain}  & {59.43}	& {42.23}	& {28.08}	& 9.40	& {23.85}	& 173.59 (2.23$\times$)
    \\ 
  & Adv-GLISTER (Ours)   &51.26	& 37.16	& 24.78	& 9.49	& 20.57	& 202.7 (1.91$\times$)
                \\ 
  &Adv-GRAD-MATCH (Ours) & 51.03	& 37.17	& 24.60	& 9.70	& 20.42	& 206.05 (1.88$\times$)
                \\ 
  &Adv-GLISTER\&Bullet (Ours)   & 53.54	& 37.24	& 23.91	& 7.69	& 20.02	&  \textbf{105.66 (3.67$\times$)}  \\ 
  &Adv-GRAD-MATCH\&Bullet (Ours) & 52.98	& 36.92	& 24.24	& 8.01	& 20.17 &  {105.61 (3.67$\times$)}
   \\ \hline
  
  \bottomrule
  \end{tabular}
  }
  \end{table*}
Recent data subset selection schemes, GRAD-MATCH \cite{pmlr-v139-killamsetty21a} and GLISTER \cite{Killamsetty_Sivasubramanian_Ramakrishnan_Iyer_2021}, have made significant contributions towards efficiently achieving high clean accuracy. We extend these approaches in the context of adversarial robustness. 
Motivated by GLISTER \cite{Killamsetty_Sivasubramanian_Ramakrishnan_Iyer_2021}, we first consider  training a subset that obtains the optimal adversarial log-likelihood  on the validation set in Eq.~\eqref{eq: glister_adv}, defined as Adv-GLISTER:
\begin{align}
{G}(\mathcal{S}) = \;\underset{ (x_V,y_V)\in \mathcal{V}} {\sum}{L_{V}}{(\theta_S; x_V+\delta_V^*, y_V}) 
\label{eq: glister_adv}
\end{align}
where $L_{V}$ is the negative log-likelihood on validation set; $\delta_V^*$ is the adversarial perturbation obtained by maximizing ${L_{V}}{(\theta_S; x_V+\delta_V, y_V})$.



Another adversarial data pruning approach is inspired by GRAD-MATCH \cite{pmlr-v139-killamsetty21a}, which aims to find the data subset whose gradients closely match those of the full training data. Adv-GRAD-MATCH is formulated as Eq.~\eqref{eq:gradmatch_adv}: 
\begin{equation}
\begin{split}
{{G}(\mathcal{S})} = \|\sum_{(x_S,y_S)\in \mathcal{S}} w \nabla_{\theta} \mathcal{L_S}(\theta;x_S+\delta_S^*, y_S) \\
-\nabla_{\theta}\underset{(x_D,y_D)\in \mathcal{D}} {\mathcal{L_D}}(\theta;x_D+\delta_D^*, y_D)\|
\label{eq:gradmatch_adv}    
\end{split}
\end{equation}

where ${w}$ is the weight vector associated with each instance $x_S$ in the subset $\mathcal{S}$; $\mathcal{L}_{S}$ and $\mathcal{L}_{D}$ denote the training loss over the subset and entire dataset; $\delta_S^*$ and $\delta_D^*$ are adversarial examples obtained by maximizing ${L_{S}}{(\theta; x_S+\delta_S, y_S})$ and ${L_{D}}{(\theta; x_D+\delta_D, y_D})$, respectively.  
During the data selection, the adversarial gradient difference between the weighted subset loss and the complete dataset loss is minimized so as to produce the optimum subset and corresponding weights.
  \begin{table*}[htbp]
  \caption{MART results where data pruning methods use only 30\% data points on CIFAR-10 and 50\% data points on CIFAR-100 for 100 epochs of training.}
  \label{tab:MART}
  \centering
  \scalebox{0.9}{
  \begin{tabular}{c|ccccccc}
  \toprule
  \hline
  \multirow{2}{*}{Dataset} & \multirow{2}{*}{Method}    & \multirow{2}{*}{Clean} & \multicolumn{3}{c}{PGD}                                      &  \multirow{2}{*}{AutoAttack} &  \multirow{2}{*}{Time/epoch (Speed-up)} \\ 
  &                         &                    & $4/255$     & $8/255$     & $16/255$    &                            &                            \\ \hline
  \multirow{6}{*}{CIFAR-10} 
  & MART \cite{Wang2020Improving} & 80.96	& 68.21	& 52.59	& 19.52	& 46.94	& 329.54 (-)    \\
  & Bullet\cite{hua2021bullettrain} & 85.29 & 70.92	& 50.64	& 13.33	& 43.77	& 199.42 (1.65$\times$)
  \\ 
  & Adv-GLISTER (Ours)   & 71.97	& 60.13	& 46.25	& 16.59	& 39.86	& 95.68 (3.44$\times$)  \\ 
  &Adv-GRAD-MATCH (Ours) & 	73.67 & 61.35 & 47.07 & 18.16 & 40.98	& 106.51 (3.09$\times$) \\ 
  &Adv-GLISTER\&Bullet (Ours)   & 73.87	& 59.89	& 44.01	& 14.20	& 38.99 & \textbf{64.31 (5.12$\times$)} \\ 
  &Adv-GRAD-MATCH\&Bullet (Ours) & 	78.78 & 64.42 & 46.72 & 13.50 & 39.53	& 77.11 (4.27$\times$)\\ \hline
  \multirow{6}{*}{CIFAR-100} 
  & MART \cite{Wang2020Improving}   & {54.85}	& {39.24}	& 25.08	& 8.59	& {22.66}	& 307.43 (-) 
     \\
  & Bullet \cite{hua2021bullettrain}  & {57.44}	& 39.22	& 24.14	& 6.66	& 21.55	& 187.73 (1.64$\times$)
    \\ 
  & Adv-GLISTER (Ours)   & 46.36	& 34.37 & 24.01	&  9.20	&19.79 	& 152.11 (2.02$\times$)
                \\ 
  &Adv-GRAD-MATCH (Ours) & 48.07	& 36.19	& {26.11}	& {10.79}	& 21.24	& 153.86 (2.00$\times$)
                \\ 
  &Adv-GLISTER\&Bullet (Ours)  & 52.13	& 35.07	& 20.67	& 5.64	& 18.21	& \textbf{100.22 (3.07$\times$)} \\ 
  &Adv-GRAD-MATCH\&Bullet (Ours) & 52.46	& 35.81	& 22.20	& 6.48	& 18.68	& 113.03 (2.72$\times$)
  
   \\ \hline
  
  \bottomrule
  \end{tabular}
  }
  \end{table*}
  \begin{table*}[htbp]
  \caption{100 {\it v.s.} 200 epoch TRADES CIFAR-10 results  with ResNet-18 when using 30\% data points with robustness regularization factor to be 1.}
  \label{tab:epoch}
  \centering
  \scalebox{0.95}{
  \begin{tabular}{ccccccccc}
  \toprule
  \hline
  \multirow{2}{*}{Method}  & \multirow{2}{*}{Epoch}  & \multirow{2}{*}{Clean} & \multicolumn{3}{c}{PGD}                                      &  \multirow{2}{*}{AutoAttack} \\  &   &   & $4/255$  & $8/255$     & $16/255$    &   &   \\ \hline
  Adv-GLISTER   & 100 & 77.62	&63.06	&46.06	&16.52	&41.61	
                \\ 
  Adv-GRAD-MATCH  & 100 & 75.61	&60.81	&45.76	&17.49	&42.19
                \\
  Adv-GLISTER    & 200 	& 78.76	&64.15	&46.11	&16.92	&42.43	
                \\ 
  Adv-GRAD-MATCH & 200	& 75.75	&61.24	&46.49	&18.55	&43.63
                \\ \hline
  
  \bottomrule
  \end{tabular}
  }
  \end{table*}
\section{ Experiments }
\subsection{Experiment Setup}
To evaluate the efficiency and generality of the proposed method, we apply   adversarial training loss functions from TRADES \cite{zhang2019theoretically} or  MART \cite{Wang2020Improving} on the standard datasets, CIFAR-10, CIFAR-100 \cite{Krizhevsky2009learning} 
trained on ResNet-18 \cite{he2016identity}. 
Our adversarial data pruning methods include Adv-GRAD-MATCH and Adv-GLISTER with different data portions (subset size) $[30\%,50\%]$ with 100 and 200 epochs where the selection interval is 20 (i.e., perform adversarial subset selection every 20 epochs of AT). The original training dataset is divided into the train (90\%) and the validation set (10\%) in Adv-GLISTER. The optimizer is SGD with momentum 0.9 and weight decay 2e-4 for TRADES and 3.5e-3 for MART. For Adv-GRAD-MATCH and Adv-GLISTER, the initial learning rate is 0.01 and 0.02 on CIFAR-10 and 0.08 and 0.05 on CIFAR-100 respectively.
    \begin{table*}[htbp]
    \caption{TRADES results on CIFAR-10 with ResNet-18 using 30\% data samples under different selection counts for 200 epoch training.}
    \scalebox{0.95}{
    \centering
    \begin{tabular}{ccccccccc}
    \toprule
    \hline
    \multirow{2}{*}{Method}  & \multirow{2}{*}{Number of selections}  & \multirow{2}{*}{Clean} & \multicolumn{3}{c}{PGD}                                      &  \multirow{2}{*}{AutoAttack} & \multirow{2}{*}{Speed-up}  \\  &   &   & $4/255$  & $8/255$     & $16/255$    &   &   \\ \hline
    TRADES \cite{zhang2019theoretically}  & - & 83.32	&68.91	&49.64	&17.31	&47.53 &-
    \\ 
    Adv-GLISTER   & 4 & 75.80	&60.48	&44.62	&16.07	&40.44 & 3.15$\times$
                  \\ 
    Adv-GRAD-MATCH  & 4 & 73.80	&60.43	&46.06	& 18.33	&43.03 &2.83$\times$
                  \\
    Adv-GLISTER    & 9 	& 78.76	&64.15	&46.11	&16.92	&42.43	 & 2.93$\times$
                  \\ 
    Adv-GRAD-MATCH & 9	& 75.75	&61.24	&46.49	&18.55	&43.63 & 2.75$\times$
                  \\ \hline
    
    \bottomrule
    \end{tabular}
    }
    \label{tab:num}
    \end{table*}
Besides the original TRADES \cite{zhang2019theoretically} and MART \cite{Wang2020Improving} methods, we also compare our approach with Bullet-Train \cite{hua2021bullettrain}. 
PGD attack \cite{madry2017towards}  ($\text{PGD-50-10}$) is adopted for evaluating the robust accuracy, ranging from low magnitude ($\epsilon=4/255$) to high magnitude ($\epsilon=16/255$) with 50 iterations as well as 10 restarts at the step-size $\alpha=2/255$ under $l_{\infty}$-norm. Moreover, AutoAttack \cite{croce2020reliable} is leveraged for the reliable robustness evaluation. Additionally, our methods can also be combined with Bullet-Train \cite{hua2021bullettrain} and we term them as Adv-GRAD-MATCH\&Bullet and Adv-GLISTER\&Bullet.







\subsection{Main Results}
\label{section:main}
Table\,\ref{tab:TRADES} shows the results of our Adv-GLISTER and Adv-GRAD-MATCH for TRADES  compared with the original TRADES and Bullet-Train methods. The comparison is in terms of clean and robust accuracy (under two attack methods, PGD Attack \cite{madry2017towards} and AutoAttack \cite{croce2020reliable}) along with the training speed-up. 
We observe that compared to the baselines, the training efficiency of our method is improved significantly on CIFAR-10, while the decrease happens on the clean accuracy and robustness under AutoAttack and PGD attacks for different values of $\epsilon$. Especially, for $\epsilon = 16/255$, the robust accuracy can be improved from $16.05\%$ (Bullet-Train \cite{hua2021bullettrain}) to $16.52\%$ and $17.49\%$ with our Adv-GRAD-MATCH and Adv-GLISTER, indicating our defensive capability on powerful attacks. 
As displayed in Table \ref{tab:TRADES}, our Adv-GRAD-MATCH and Adv-GLISTER reduce the training overheads (seconds per epoch) enormously and achieve $3.44\times$  and $3.09\times$ training speed-ups. 
After combining our approaches with Bullet-Train \cite{hua2021bullettrain}, an even faster acceleration of $5.12\times$ can be reached. 

On CIFAR-100, the validity of our schemes is consistent as well. The reason why both clean and robust accuracy drop might be that our data pruning schemes struggle with the dimensionality and complexity of the dataset. Regardless, our schemes still result in conspicuous computation savings compared with other baselines.


To understand the robustness improvements of our schemes, we  track the dynamics of the outlier, robust, and boundary sets (similar to \cite{hua2021bullettrain}) using $\text{PGD-5-1}$ attack. Without any attack, the outlier examples have already been mistaken by the model, but boundary and robust examples are correctly identified. After adversarial attacks, boundary examples are incorrectly classified while robust examples are still correctly classified. Fig.\,\ref{fig:tracking} displays the dynamics of the outlier, boundary, and robust examples on CIFAR-10 for various schemes. During the model training and data selection, the number of robust samples gradually increases and   eventually   dominates, while the number of outliers and boundary data points decreases over epochs, revealing similar achievements in TRADES-based AT and data pruning-based methods. In addition, the ultimate portions of three sets explain the clean accuracy and robustness degrading of our approaches. In detail, two baselines obtain more robust samples and fewer boundary and outlier examples. 

We further evaluate the performances of adversarial data pruning based on the loss of MART in Table\,\ref{tab:MART}. Results are consistent with our findings on TRADES in Table\,\ref{tab:TRADES}.

\subsection{Ablation Studies}
\textbf{\emph{Epoch.}} 
We first consider the training epoch. Table\,\ref{tab:epoch} shows that longer training improves both clean and robust accuracy. Due to the shrinking data size, more epochs are required to enhance data-efficient adversarial learning, in alignment with standard data pruning training. However, 100-epoch training appears to be sufficient for the small dataset.

\begin{figure*}[h]
\centering
\subfigure[TRADES.]{\includegraphics[width=0.496\linewidth]{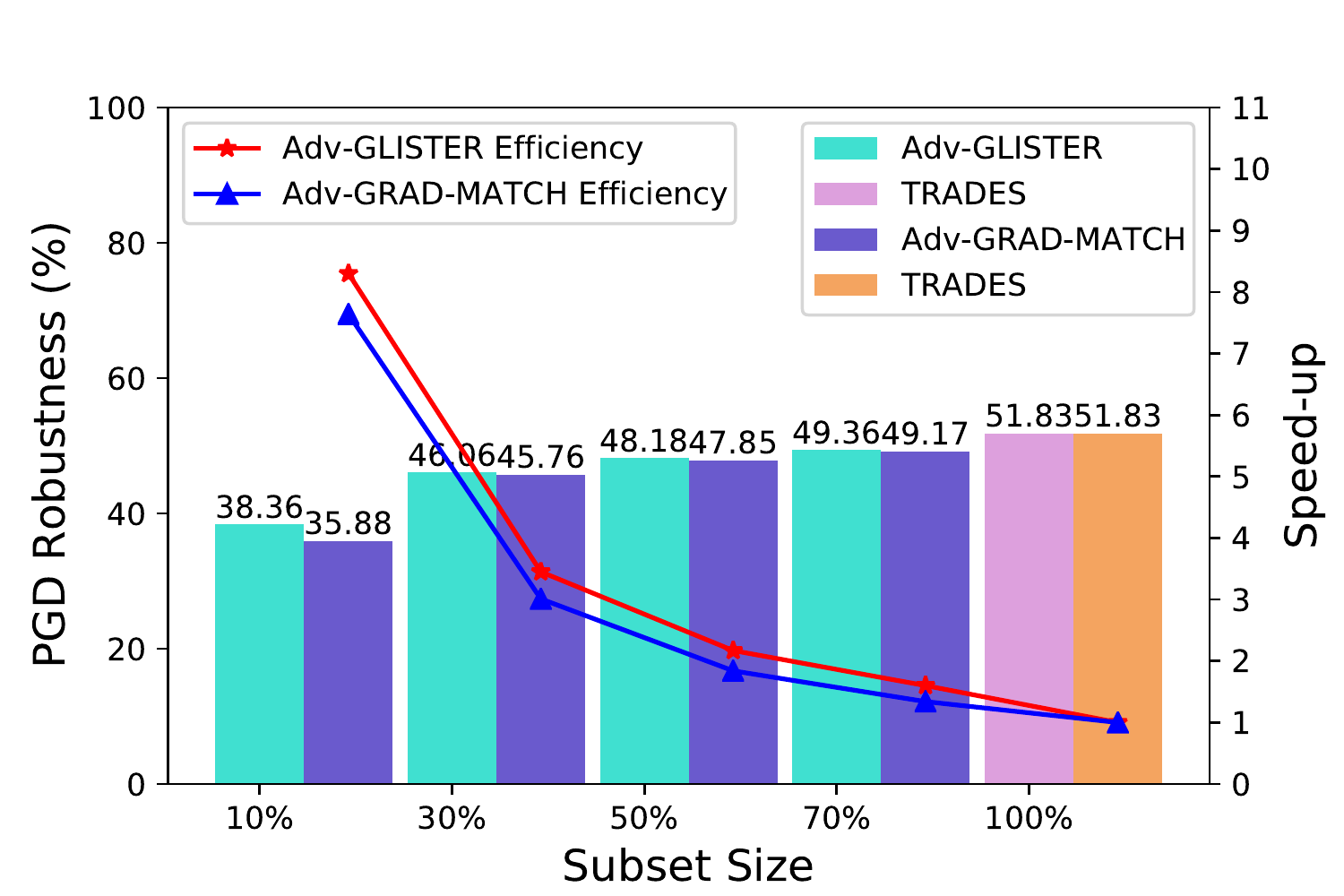}}\vspace{-1.8mm}
\subfigure[MART.]{\includegraphics[width=0.496\linewidth]{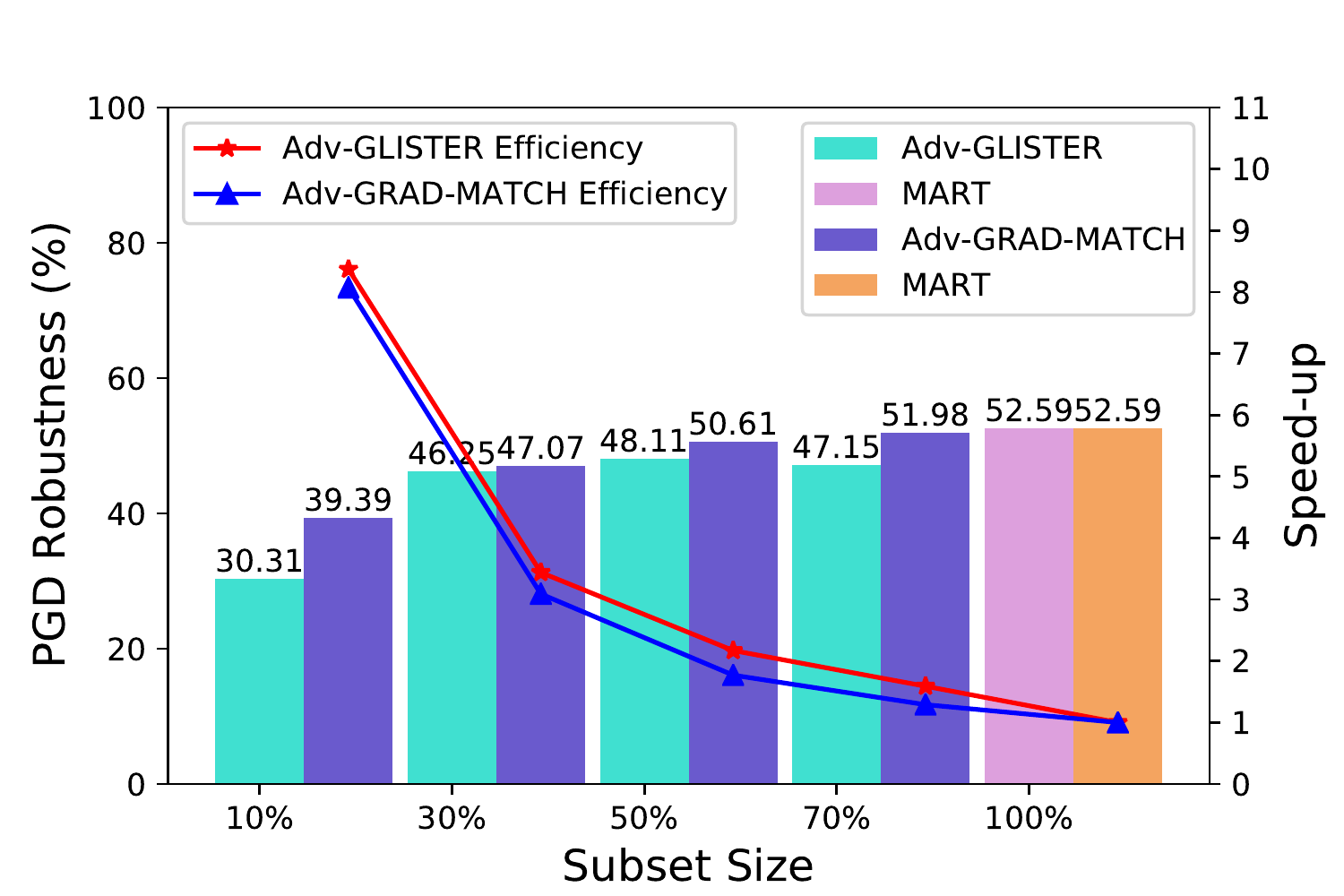}}\vspace{-1.8mm}

\caption{\footnotesize{PGD evaluation {($\epsilon = 8/255$)}  with the corresponding speed-up under different subset sizes for 100 epoch CIFAR-10 training. Note that when the size is 100\%, data pruning methods are not applied and the speed-up is compared with the baselines (TRADES or MART). }}

  \label{fig:portion}  
\end{figure*}

\textbf{\emph{Subset Size.}} 
We experiment with different subset sizes. Moving from the extremely small subset (10\% of the full training set) to a larger subset (70\%) in Fig.\,\ref{fig:portion}, the observation is that robust accuracy gradually increases  to that of the full dataset. This highlights the benefit of pruning with optimal subset size. We can see that 30\% is an appropriate choice for the CIFAR-10 subset size, after taking the global efficiency into account.

\textbf{\emph{Number of selection rounds.}} 
{In Sec.\,\ref{section:main}}, our experiments perform adversarial data pruning every 20 epochs (with 9 selections). Here we present the results of data pruning every 40 epochs (with 4 selections). As shown in Table \ref{tab:num}, 9 selections can achieve better clean and robust accuracy with comparable acceleration. 

\section{ Conclusion and Future Work }
In this paper, we investigated efficient adversarial training from a data-pruning perspective. 
With comprehensive experiments, we demonstrated that proposed adversarial data pruning approaches outperform the existing baselines by mitigating substantial computational overhead.  These positive results pave a path for future research on accelerating AT by minimizing redundancy at the data level. 
Our future work will focus on designing more accurate pruning schemes for large-scale datasets. 

\section*{Acknowledgment}
This work was performed under the auspices of the U.S. Department of Energy by Lawrence Livermore National Laboratory under Contract DE-AC52-07NA27344 and was supported by LLNL-LDRD Program under Project No. 20-SI-005 (LLNL-CONF-842760). 

\bibliography{sample-ceur}

\appendix



\end{document}